\documentclass{article} 

\usepackage{algorithm}
\usepackage{algorithmic}
\usepackage{amsmath, amssymb}
\usepackage{bm}
\usepackage{booktabs}
\usepackage{dsfont}
\usepackage{graphicx}
\usepackage{hyperref}
\usepackage{mathtools}
\usepackage{multirow}
\usepackage{natbib}
\usepackage[separate-uncertainty=true,multi-part-units=repeat]{siunitx}
\usepackage{xcolor}
\usepackage{xspace}

\usepackage{adjustbox}

\newcommand{\ours}{ADRA\xspace}
\newcommand{\B}{\bfseries}

\definecolor{blue}{HTML}{2778B1}
\newcommand{\bluebullet}{\textcolor{blue}{\textbullet}}
\definecolor{green}{HTML}{3AA039}
\newcommand{\greenbullet}{\textcolor{green}{\textbullet}}
\definecolor{orange}{HTML}{FC8126}
\newcommand{\orangebullet}{\textcolor{orange}{\textbullet}}
\definecolor{purple}{HTML}{6A3997}

\definecolor{red}{HTML}{E02025}
\newcommand{\redbullet}{\textcolor{red}{\textbullet}}

\makeatletter
\DeclareRobustCommand\onedot{\futurelet\@let@token\@onedot}
\def\@onedot{\ifx\@let@token.\else.\null\fi\xspace}
\def\eg{\emph{e.g}\onedot} 
\def\ie{\emph{i.e}\onedot} 
\def\cf{\emph{c.f}\onedot} 
\def\etc{\emph{etc}\onedot}

\makeatother

\usepackage{hyperref}
\usepackage{url}

\title{Deep Anomaly Detection by Residual Adaptation}

\author{Lucas Deecke \\
University of Edinburgh \\
\texttt{l.deecke@ed.ac.uk} \vspace*{.2cm} \\ 
Lukas Ruff \\
Technische Universit{\"a}t Berlin \\
\texttt{lukas.ruff@tu-berlin.de} \vspace*{.2cm} \\
Robert A. Vandermeulen \\
Technische Universit{\"a}t Berlin \\
\texttt{vandermeulen@tu-berlin.de} \vspace*{.2cm} \\
Hakan Bilen \\
University of Edinburgh \\
\texttt{hbilen@ed.ac.uk}}

\begin{document}

\maketitle

\begin{abstract}

Deep anomaly detection is a difficult task since, in high dimensions, it is hard to completely characterize a notion of ``differentness'' when given only examples of normality. In this paper we propose a novel approach to deep anomaly detection based on augmenting large pretrained networks with residual corrections that adjusts them to the task of anomaly detection. Our method gives rise to a highly parameter-efficient learning mechanism, enhances disentanglement of representations in the pretrained model, and outperforms all existing anomaly detection methods including other baselines utilizing pretrained networks. On the CIFAR-10 one-versus-rest benchmark, for example, our technique raises the state of the art from 96.1 to 99.0 mean AUC. 

\end{abstract}

\section{Introduction}\label{sec:introduction}

The core goal of anomaly detection is the identification of unusual samples within data \citep{edgeworth1887,grubbs1969,schoelkopf99,chandola09}. What complicates matters is that unusualness can be caused by a variety of factors, especially for data types that are semantically rich. For these settings, there has been continued interest in developing new deep anomaly detectors \citep{zhai2016,schlegl17,sabokrou2018,deecke18,ruff18,golan18,pidhorskyi2018,hendrycks18,hendrycks19b} that utilize end-to-end learning, a defining property amongst deep learning approaches \citep{krizhevsky12,he16}.

For deep anomaly detection there is no natural learning objective, and thus several methods have been proposed. One emerging trend is to utilize self-supervision \citep{golan18,hendrycks19b,bergman20a}. In these approaches one creates an auxiliary task from a nominal dataset by transforming its samples, and then utilizing these in a fashion that resembles supervision. A different approach uses large unstructured collections of data, which serve a purpose similar to outliers \cite{hendrycks18}, to train models akin to classifiers \citep{ruff20a}, or enhance self-supervised learning criterions \citep{hendrycks19b}. Considering the rather simplistic objective for these approaches, especially compared to the richness of images, one may wonder whether they learn particularly meaningful features from such training procedures. This is potentially problematic since anomalies can manifest themselves in subtle ways that require a good semantic understanding: for example, anomalous objects may appear in crowded scenes \citep{mahadevan10}, or be subject to transitions between day and night in video footage \citep{sultani18}. 

Recently there has been a large increase in the availability and utilization of pretrained networks. Ideally, these will enhance the representations learned by task-specific downstream models, as they already incorporate different variations commonly seen in data (edges, color, semantic parts, \etc). While \citet{he19} fundamentally questioned whether actual benefits are brought about by pretrained models, \citet{hendrycks19a} recently painted them in a more positive light, showing they boost performance in robustness and uncertainty tasks.

Pretrained representations are relied upon in many areas of machine learning, for example in object detection \citep{girshick14,girshick15}, transfer learning \citep{guo19}, when looking to transfer between large numbers of tasks \citep{zamir18}, or from one domain to another \citep{rebuffi17,rebuffi18}. Other examples can be found in natural language processing, where a surge of papers has recently elevated the role of pretrained models \citep{mikolov18,devlin18,howard18,adhikari19,beltagy19,hendrycks20}. 

While anomaly detection fundamentally differs from common pretraining tasks (\eg ImageNet \citep{deng09}) since it is more abstract and less well-defined, it may still benefit from adapting rich, pretrained representations. In doing so, one should however ensure that the change in representation is not excessive \citep{li18}, as this risks \emph{catastrophic forgetting} \citep{kirkpatrick17}. For anomaly detection in particular, it is crucial to preserve variations incorporated during pretraining that, even though they potentially don't exist in the training set, can nonetheless be meaningful for inferring anomalous characteristics at test time. On the other hand, it is important to let the network have \emph{some} flexibility to learn new variations which are important for the new task.

\begin{figure}[t]
\begin{center}
\centerline{\includegraphics[width=.8\textwidth]{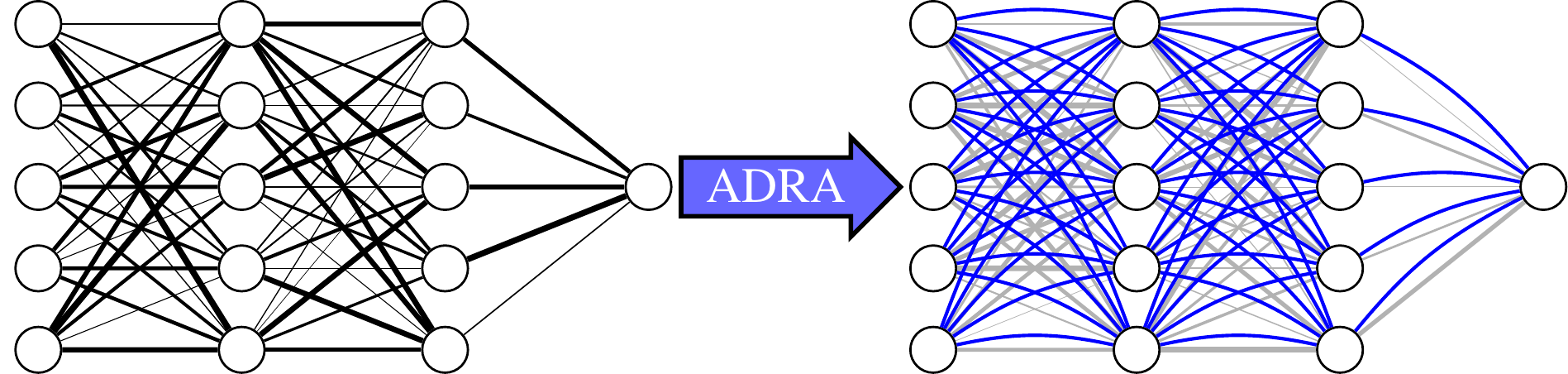}}
\end{center}
\caption{Instead of finetuning parameters of pretrained models, \ours keeps them fixed and injects new learnable connections into the network (symbolized in blue). Extending the model with new parameters lets \ours incorporate representations suitable for deep anomaly detection, while holding on to information incorporated in the pretraining task. Best viewed in color.} \label{fig:sketch}
\end{figure}

In this paper we introduce an anomaly detection method that goes beyond the traditional finetuning paradigm by using lightweight dynamic enhancements \citep{deecke20}, which serve as modifications to the pretrained network at every layer; for a visualization, see Figure~\ref{fig:sketch}. We call this \emph{anomaly detection with residual adaptation} (ADRA). This introduces a simple enhanced objective that combines outlier exposure \citep{hendrycks18} and deep one-class classification \citep{ruff20b}, two powerful learning techniques for anomaly detection. \ours is straightforward to train and deploy, highly parameter-efficient, and can much better consolidate pretrained networks and anomaly detection than mere feature extraction \citep{bergman20b}. In extensive experiments we show that \ours outperforms all previous approaches in the deep anomaly detection literature on a set of common benchmarks. On the CIFAR-10 one-versus-rest benchmark, for example, our technique raises the state of the art from 96.1 to 99.0 mean AUC, reducing the gap to perfect performance by 75\%. Besides the strong performance of \ours, we use images from a new disentanglement dataset \citep{gondal19} to show that \ours naturally disentangles meaningful variations in data into its representations.

\section{Related work}
\label{sec:background}

Anomaly detection has a long history, with early work going back to \citet{edgeworth1887} and has been extensively studied in the classical machine learning literature, \eg through generative models for intrusion detection \citep{yeung02}, or hidden Markov models for registering network attacks \citep{ourston03}. Other examples include active learning of anomalies \citep{pelleg05}, or dynamic Bayesian networks for traffic incident detection \citep{singliar06}. An overview of traditional anomaly detection methods can be found in \citet{chandola09}, an empirical evaluation in \citet{emmott13}. 

Previous deep anomaly detection methods utilized autoencoders \citep{zhou17,zong18}, hybrid methods \citep{erfani16}, or generative adversarial networks \citep{schlegl17,akcay18,deecke18,perera19}. A recent focus is on repurposing auxiliary tasks for anomaly detection, often following the paradigm of self-supervision: \citet{golan18} propose learning features from predicting geometric transformations of the nominal data, which was extended to other data types by \citet{bergman20a}. In a separate line of work, \citet{hendrycks18} propose carrying out anomaly detection through a paradigm they call \emph{outlier exposure} where one utilizes large unstructured sets of data, assumed to not belong to the normal class, to improve performance of deep anomaly detection. Our approach also leverages learning from such corpora, however bypasses all self-supervision steps entirely.

The technique of adding residual connections to adapt networks to new tasks, also known as \emph{residual adaptation}, was introduced in \citet{rebuffi17,rebuffi18}. While originally developed for multi-task learning, this assumption was recently relaxed in work that extended residual adaptation to other problem settings, such as latent domain learning \citep{deecke20}. In the realm of language modeling \citet{stickland19} applied residual adaptations to pretrained BERT networks \citep{devlin18} to improve performance there. Our method further demonstrates the usefulness of residual adaptation outside of multi-task learning by extending it to the task of anomaly detection.

A number of recent publications proposed unsupervised mechanisms to learn disentangled representations \citep{kulkarni15,higgins17,bouchacourt18,burgess18,chen18,kim18,kumar18}. \citet{locatello19} outlined the incompatibility of unsupervised learning and repesentation disentanglement, and follow-up work established the need for some form of weak supervision to give rise to disentanglement \citep{locatello20}. Considerable hopes have been placed on the usefulness of such disentangled representations \citep{bengio17,steenkiste19}. We investigate connections to anomaly detection in Section \ref{sec:disentanglement}.

\section{Method}
\label{sec:method}

We review the individual components to our proposed approach in Sections \ref{sec:hsc} and \ref{sec:ra}, then subsequently introduce \ours in Section \ref{sec:adra}.

\subsection{Deep one-class classification}
\label{sec:hsc}

When learning from data, a semantic understanding of normality is typically extracted from a set of data $S_n = \{x_j\}_{j=1}^n$ assumed to have been sampled i.i.d.~from the nominal distribution $\mathbb P$ over some sample space $\mathcal X$. How this data is then incorporated is how the different approaches in the anomaly detection literature can be categorized, \eg in an unsupervised way \citep{ruff18}, or through self-supervision \citep{golan18}, \cf Section \ref{sec:background}. The ansatz of outlier exposure \citep{hendrycks18} revolves around the utilization of a large number of unlabeled images from some unstructured corpus of data $Q_m$ (where potentially $m \gg n$), for example 80 Million Tiny Images \citep{torralba08}, on which models are trained to identify whether samples belong to the corpus, or the nominal data. Importantly, this is a form of weak supervision via existing resources \citep{zhou18}, and not equivalent to binary classification: images from the corpus are not necessarily outliers (and may even contain samples from $\mathbb P$). Nonetheless, this procedure can help models incorporate richer representations of the data in $S_n$.

Initially observed in \citet{ruff20a}, the structure of anomaly detection tasks benefits from encapsulating the normal class through radial functions, in line with the so-called \emph{concentration assumption} fundamental in anomaly detection \citep{scholkopf02,steinwart05}. \citet{ruff20b} showed that outlier exposure also benefits from a reformulation via a class of spherical learning objectives, which the authors use to set the current state of the art in anomaly detection performance. Given access to data, this learning criterion can be expressed as a functional of some model $f$ as

\begin{equation}\label{eq:hsc}
\mathcal H[f] = \frac1{N} \sum_{i=1}^{N} y_i h(f(x_i)) + (1-y_i) \log \{ 1 - \exp(-h(f(x_i)))\},
\end{equation}

where pseudo-labels $y_i$ are determined by a sample's origin, \ie $y(x_i) = \mathds 1_{x_i \in Q_m}$. This loss can be coupled with different radial functions $h$, which \citet{ruff20b} recommend setting to $h(f(x)) = \sqrt{\Vert f(x) \Vert^2 + 1} - 1$. We follow their recommendation in this work, and found it to be stable across experiments.

Note that previous works (to which we compare in our experiments, see Section \ref{sec:experiments}) would use some randomly initialized neural network and obtain its parameterization via minimization of some criterion, in the above case for example $\theta = \arg\min_{\theta'} \mathcal H [f_{\theta'}]$. In \ours, we constrain the optimization to a more suitably regularized class of functions, see Section \ref{sec:adra}.

\subsection{Residual adaptation}
\label{sec:ra}

In conventional residual networks \citep{he16}, the information from the $l$'th layer is passed on via $x_{l+1} = x_l + f_{\theta_l}(x_l)$, where each $f_{\theta_l}$ is typically parameterized as a 3x3 convolution.\footnote{This omits normalization and activation to declutter notation.} Originally developed in the context of multi-domain learning, \citet{rebuffi18} proposed adding a small linear correction to every layer, such that
\begin{equation}
x_{l+1} = x_l + f_{\theta_l}(x) + h_{\alpha_l}(x),
\end{equation}

with each $h_{\alpha_l}$ parameterized by a smaller 1x1 convolution. \citet{deecke20} recently generalized this concept through a mixture of experts approach. For this, a set of $K$ linear corrections $\{h_{\alpha_k}\}_{k=1,\dots,K}$ is introduced at every layer, which are targeted via a self-attention \citep{lin17} mechanism $g_k$ that adaptively combines available corrections. This yields
\begin{equation}
x_{l+1} = x_l + f_{\theta_l}(x) + \sum_{k=1}^K g_{lk}(x) h_{\alpha_{lk}}(x).
\end{equation}

As the authors show, residual adaptation gives rise to an efficient plug-in module that encourages parameter sharing between similar modes in data. Crucially, the module also increases the robustness of models in regions where the density associated with the data-generating distribution has little mass\,---\,regions of particular importance to anomaly detection.

The motivation behind residual adaptation was developed in earlier work on universal representations \citep{bilen17}. At their core, universal representations build on the idea that general-purpose parameters obtained through some large pretraining task require only small modifications for them to be adapted to a wide range of tasks. Even though universal representations were originally conceived with the objective of training compact models over sets of very different tasks, our experiments suggest that its insights hold promise for a wider range of learning problems, anomaly detection included.

\subsection{Anomaly detection with residual adaptation}
\label{sec:adra}

Following work that investigated the prospects of large pretrained networks \citep{devlin18,howard18,adhikari19,beltagy19,hendrycks20}, a recent study proposed carrying out anomaly detection through a nearest neighbor search on top of features extracted from a large pretrained residual network \citep{bergman20b}. We include this approach in our experiments (see Table~\ref{table:cifar-10}), but as its performance shows, simply transferring over fixed representations to an unrelated task does not sufficiently incorporate abstract information, limiting its usefulness for complex, high-dimensional anomaly detection. Next, we outline how parameters obtained from pretraining can be more adequately repurposed for a new task.

The initial pretraining itself follows a simple protocol: a model's parameters are randomly initialized from some distribution over parameters (for example Xavier initialization \citep{glorot10}). Minimization of a suitable pretraining task $\mathcal A$ (say, object classification on ImageNet \citep{deng09}) then yields a set of general-purpose parameters $\theta$. The so-obtained model $f_\theta$ is now fully ``pretrained'', ready to be used in a downstream task $\mathcal B$.

The traditional ansatz for leveraging pretrained models is to ``finetune'', \ie continuing to optimize the model parameters (or a subset thereof) on $\mathcal{B}$. Finetuning can therefore be seen as a type of weight initialization, using weights from a pretrained network. One crucial limitation of this learning protocol is that when learning on $\mathcal B$ isn't carried out very carefully through the introduction of some explicit inductive bias \citep{li18}, this risks \emph{catastrophic forgetting} of information previously extracted from $\mathcal A$. To alleviate this a common approach is to develop adequate forms of regularization, which were used successfully \eg in continual learning \citep{kirkpatrick17,lopezpaz17}. The explicit bias we introduce in \ours is different, and instead obtained by sidestepping the standard finetuning protocol and directly modifying the model's structure.

To modulate the base network, \ours introduces a fresh set of model parameters $\alpha \sim \mathbb P_\alpha$. These have not previously been exposed to the pretraining task $\mathcal A$, and serve to incorporate the task-specific variations crucial to inferring anomaly. At the same time, \ours fixes the pretrained parameters and never changes them, which ensures they remain linked to the information obtained in $\mathcal A$. In doing so, our methodology constraints the learning criterion, recasting the optimization of the radial loss functional introduced in eq.~\eqref{eq:hsc} as

\begin{equation}\label{eq:adra}
f_{\theta\alpha} = \underset{\alpha'}{\arg\min} \, \mathcal H[f_{\theta\alpha'}], \,\,\,\,\,\,\, \text{s.t.} \,\theta=\underset{\theta'}{\arg\min} \, \mathcal A[f_{\theta'}].
\end{equation}

Because the pretrained parameters $\theta$ are fully determined via $\mathcal A$ and fixed thereafter, only the new set of residual connections $\alpha$ is task-specific. For every nominal class, \ours therefore only requires the parameters specific to $\mathbb P$ to be inserted back into the model. This requires a much smaller number of overall model parameters (\ie $\vert \alpha \vert \ll \vert \theta \vert$), giving rise to a highly efficient, adaptive architecture $f_{\theta\alpha}$.

\section{Experiments}
\label{sec:experiments}

We consider three settings to evaluate our method: anomaly detection on the (i.)~one-versus-rest and (ii.)~hold-one-out benchmarks, as well as (iii.) measuring the disentanglement of learned representations. All experiments use code implemented through standard routines in PyTorch \citep{paszke17}.

\subsection{One-versus-rest anomaly detection}
\label{ssec:one-versus-rest}

We evaluate performance of \ours on the CIFAR-10 \citep{krizhevsky09} one-versus-rest anomaly detection benchmark, which is reported across large parts of the literature \citep{deecke18,golan18,hendrycks18,ruff18,abati19,hendrycks19b,perera19,bergman20a,ruff20a,ruff20b}. This benchmark is not equivalent to CIFAR-10 classification, and consists of ten individual tasks instead: in each, a single class $y_c$ is fixed as the normal class\,---\,say, dogs. All dogs in the CIFAR-10 training split are collected into $S_n$, from which models can then be learned about the nominal distribution $\mathbb P_{y_c}$. Finally, models are evaluated against the entire CIFAR-10 test split, and performance is recorded by checking whether anomaly scores assigned to dogs are lower than scores assigned to the remaining non-dog classes. To express performance through a single number, authors usually report AUC under the receiver operating characteristic; we follow this practice here.

In \ours, we additionally contrast $S_n$ against images from an unstructured corpus $Q_m$. Guided by previous work \citep{hendrycks18,ruff20b}, we fix this to contain all samples from the CIFAR-100 training split. Access to $Q_m$ should best be thought of as a weak supervisory signal: CIFAR-100 does not contain any of the classes in CIFAR-10, so 9 out of 10 classes are only seen during evaluation, making them true outliers.

\paragraph{Optimization}

Training is carried out using stochastic gradient descent (momentum parameter of 0.9, weight decay of \num{e-4}) for a total of 120 epochs, with learning rate reductions by 1/10 after 80 and 100 epochs. The batch size is fixed to 128, with each batch containing an equivalent number of samples from $S_n$ and $Q_m$. All experiments use a residual network with 26 layers, and unless otherwise noted, its initial parameters were obtained from pretraining on a downsized ImageNet variant (at 72x72 resolution), for a final top-1 accuracy of 60.32\%. \ours parameters $\alpha$ are always initialized randomly, and then trained on the constrained objective in eq.~\eqref{eq:adra}. We vary the number of linear corrections in \ours over $K=\{1,2,4\}$. Performances for \ours are recorded by averaging over five random initializations.

\paragraph{Results}

\begin{table}[t]
\centering
\caption{AUCs for different methods on the CIFAR-10 one-versus-rest anomaly detection benchmark. For \ours we vary $K=\{1,2,4\}$.}\label{table:cifar-10}
\vspace*{.4cm}
\resizebox{\columnwidth}{!}{
\begin{tabular}{ccccccccc}
\toprule
         &      &      &      &      &      & \multicolumn{3}{c}{\ours}            \\   
Class    & GT 
         & IT 
         & kNN-AD 
         & GT+ 
         & SAD 
         & $K\!=1$
         & $K\!=2$
         & $K=4$       \\
\midrule
0        & 74.7 & 78.5 & 93.9 & 90.4 & 96.4 & 98.8       & 99.0       &\B99.1      \\
1        & 95.7 & 89.8 & 97.7 & 99.3 & 98.8 & 99.7       &\B99.8      &\B99.8      \\
2        & 78.1 & 86.1 & 85.5 & 93.7 & 93.0 & 96.3       & 98.1       &\B98.6      \\
3        & 72.4 & 77.4 & 85.5 & 88.1 & 90.0 & 95.7       & 96.3       &\B97.3      \\
4        & 87.8 & 90.5 & 93.6 & 97.4 & 97.1 & 98.2       &\B99.1      & 99.1       \\
5        & 87.8 & 84.5 & 91.3 & 94.3 & 94.2 & 97.4       & 98.1       &\B98.2      \\
6        & 83.4 & 89.2 & 94.3 & 97.1 & 98.0 & 99.4       &\B99.6      &\B99.6      \\
7        & 95.5 & 92.9 & 93.6 & 98.8 & 97.6 & 99.1       &\B99.5      &\B99.5      \\
8        & 93.3 & 92.0 & 95.1 & 98.7 & 98.1 & 99.4       & 99.4       &\B99.5      \\
9        & 91.3 & 85.5 & 95.3 & 98.5 & 97.7 & 99.1       &\B99.4      & 99.3       \\
\midrule
Mean AUC & 86.0 & 86.6 & 92.5 & 95.6 & 96.1 & 98.3       & 98.8       &\B99.0      \\
\bottomrule
\end{tabular}
}
\end{table}

As shown in Table~\ref{table:cifar-10}, \ours raises the state of the art to 99.0 mean AUC, closing the gap between the previous best method of 96.1 mean AUC (SAD \citep{ruff20b}) and perfect classification by roughly 75\%. As demonstrated by the performance of kNN-AD \citep{bergman20b}, a different work also focusing on the utilization of pretrained networks for anomaly detection, simply using features from a large pretrained network alone does not solve the problem of detecting anomalies. Our results suggest including adaptation in a proper way is critical for utilizing these networks to their full potential. 

Dynamic residual adaptation with self-attention as proposed by \citet{deecke20} improves performance across the benchmark. For larger $K$, the improvement in effectiveness is reduced, as sufficient adaptivity is likely already obtained for a smaller number of experts. Unless otherwise noted, in subsequent trials we therefore fix $K\!=2$.

Current existing methods require that all parameters of each model are stored. This scales their memory requirements as $\mathcal O(T\vert \theta \vert)$, where $T$ denotes the number of subtasks (\eg ten for our benchmarks). For \ours, a small set of task-specific corrections augment the base model, reducing its computational footprint to $\mathcal O(\vert \theta \vert + KT\vert \alpha \vert)$. On our benchmarks for example, \ours with $K=2$ requires only around \num{20.2}{mil} parameters in total, a fraction of the roughly \num{62.1}{mil} parameters required to parameterize ten individual ResNet26. We visualize these savings in Figure~\ref{fig:disentanglement} (left).

\subsection{Robustness to small modes}

Ideally models have the ability to incorporate information from nominal samples even if they form only a minor mode of $\mathbb P$, such that only few samples from this mode are contained in $S_n$. \citet{deecke20} suggest that dynamic residual adaptation benefits the robustness to small latent domains, \ie regions where the density associated with the data-generating distribution has little mass. Next, we evaluate this property in the context of anomaly detection.

For this experiment, we let the normal class be constituted by samples associated with pairs of classes $(y_a,y_b)$, such that $S_n \sim \mathbb P_{y_a} + r \mathbb P_{y_b}$, with $r\in [0,1]$ controlling the presence of samples from $y_b$. For a robust model, even as $\mathbb P$ is relaxed toward $\mathbb P_{y_a}$, its ability to detect normality amongst the secondary class remains intact.

\begin{figure}[t]
\begin{center}
\centerline{\includegraphics[width=\textwidth]{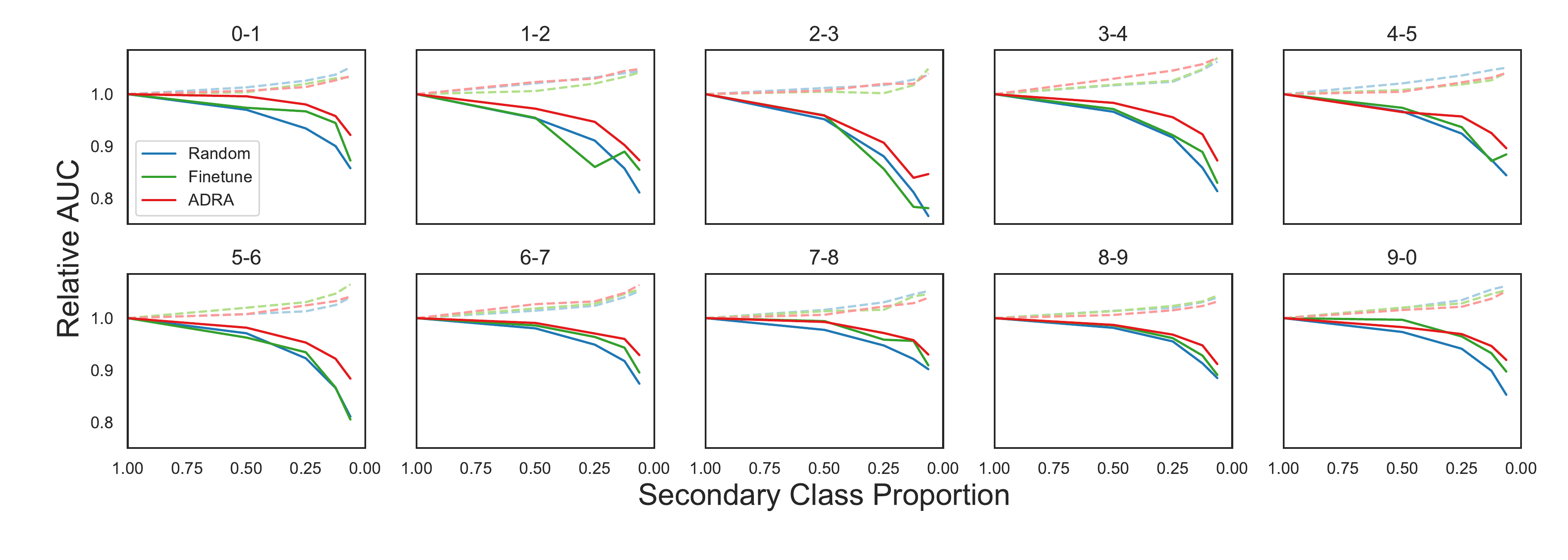}}
\end{center}
\vspace*{-.8cm}
\caption{Relative AUCs for the secondary class for pairs of classes from CIFAR-10. Shown are performances for a random initialization (\bluebullet), traditional finetuning (\orangebullet), and \ours (\redbullet). Dashed curves indicate the relative change in performance for the primary class.} \label{fig:secondary}
\end{figure}

We pair up classes from CIFAR-10, and report AUCs relative to $r=1$ for the primary and secondary class in Figure~\ref{fig:secondary}. When models are trained from a random initialization, their performance falls of faster than for approaches that start from pretrained networks. This trend is consistent across class pairings. There is a modest increase in performance for traditional finetuning, \ours offers the highest robustness at incorporating small nominal modes.

\subsection{Hold-one-out anomaly detection}
\label{ssec:hold-one-out}

An alternative benchmark implemented in \citet{perera19} or \citet{ahmed19} is to collect multiple classes into the nominal data $S_n$, while only a single class is declared anomalous and set aside during training. At test time then, the task is to identify samples from the held-out class. It has been argued that this is a more difficult benchmark than one-versus-rest, since it requires the learning of multiple nominal modes \citep{ahmed19,bergman20b}.

On the hold-one-out benchmark, \citet{ahmed19} evaluate the performance of ranking anomaly via maximum softmax probability \citep{hendrycks17} and ODIN \citep{liang18} in combination with an auxiliary self-supervised criterion inspired by RotNet \citep{gidaris18}. In particular, the authors propose evaluating models on STL-10 \citep{coates11} for a novel, more difficult benchmark: its images have a higher resolution of 96x96, while only containing 500 samples for each object class. The dataset also contains a large unlabeled split, which we collect into our corpus $Q_m$ for outlier exposure.

For STL-10, we pretrained ResNet26 for a final top-1 classification accuracy of 63.74\% on ImageNet at this resolution. All other optimization settings remain unchanged from those outlined in Section \ref{ssec:one-versus-rest}.

\begin{table}[t]
\centering
\caption{Average precisions on CIFAR-10 and STL-10 hold-one-out benchmarks. \ours uses $K\!=2$.}\label{table:stl-10}
\vspace*{.4cm}
\resizebox{\columnwidth}{!}{
\begin{tabular}{ccccccccc}
\toprule
               & \multicolumn{4}{c}{CIFAR-10}   & \multicolumn{4}{c}{STL-10}                                         \\
\cmidrule(l){2-5}
\cmidrule(l){6-9}
Class          & RA-ODIN 
               & TF
               & L\textsuperscript{2}-SP 
               & \ours
               & RA-ODIN 
               & TF
               & L\textsuperscript{2}-SP 
               & \ours \\
\midrule
0              &\B49.8     & 43.0     & 50.8     & 42.9    & 23.4     & 23.1                &\B44.5                   &  43.1   \\
1              & 17.4      & 76.7     & 77.4     &\B88.4   &\B40.1    & 13.8                & 14.8                    &  23.7   \\
2              & 54.6      & 61.1     & 68.0     &\B74.4   & 16.9     & 39.9                &\B82.2                   &  56.0   \\
3              & 55.8      & 65.8     &\B72.7    & 72.5    & 31.4     & 18.9                & 27.4                    & \B37.5  \\
4              & 52.8      & 60.6     & 65.3     &\B73.3   &\B29.7    & 25.3                & 17.0                    &  28.8   \\
5              & 32.5      & 64.2     &\B65.1    & 63.3    &\B26.1    & 17.3                & 12.3                    &  18.5   \\
6              & 54.4      & 84.0     & 89.9     &\B90.7   & 23.6     & 30.1                & 22.1                    & \B33.5  \\
7              & 39.7      & 52.9     & 46.8     &\B53.2   & 28.3     & 18.4                & 14.9                    & \B44.2  \\
8              & 28.8      & 70.8     & 64.2     &\B74.4   & 15.4     & 49.2                &\B70.3                   &  59.4   \\
9              & 29.9      & 87.7     & 84.5     &\B94.1   & 16.6     & 40.7                & 44.4                    & \B55.7  \\
\midrule
Mean AP        & 41.2      & 66.7     & 68.5     &\B72.7   & 25.1     & 27.7                & 35.0                    & \B40.0  \\
\bottomrule
\end{tabular}
}
\end{table}

\paragraph{Results}

We follow \citet{ahmed19} and report performance on the hold-one-out benchmark in terms of average precision. We include their best results, rotation-augmented ODIN (RA-ODIN), in Table~\ref{table:stl-10}. We do not include the results for OCGAN from \citet{perera19} in our tabulation, as the authors report in AUC instead. Note however \ours outperforms OCGAN on mean AUC by a wide margin: 94.7 versus 65.7.

As our results confirm, inferring anomaly on STL-10 is significantly harder. In particular, traditional finetuning (TF) does not successfully address the task of anomaly detection, likely because variations that are important to determining anomaly at test time are overwritten in the finetuning process, yielding poor performance across classes (mean AP of 27.7). To counteract this, we follow the recommendation of \citet{li18} and add a penalty toward the initial pretrained parameters via L\textsuperscript{2}-SP regularization, scaled with a regularization strength of $\alpha=\num{e-2}$. While this boosts performance somewhat (mean AP of 35.0), simply regularizing the otherwise unchanged finetuning process does not fully address the issue of forgetting information extracted during pretraining.

By modifying the structure of the base model, \ours introduces a different explicit bias toward the initial pretrained representation. Our results indicate that preserving the initial pretrained parameters and adapting the network to the task through the introduction of new network components allows one to keep both the benefits of pretraining and learning for a specific task.

\subsection{Disentanglement of representations}
\label{sec:disentanglement}

\begin{figure}[ht]
\begin{center}
\centerline{
\includegraphics[width=.47\textwidth]{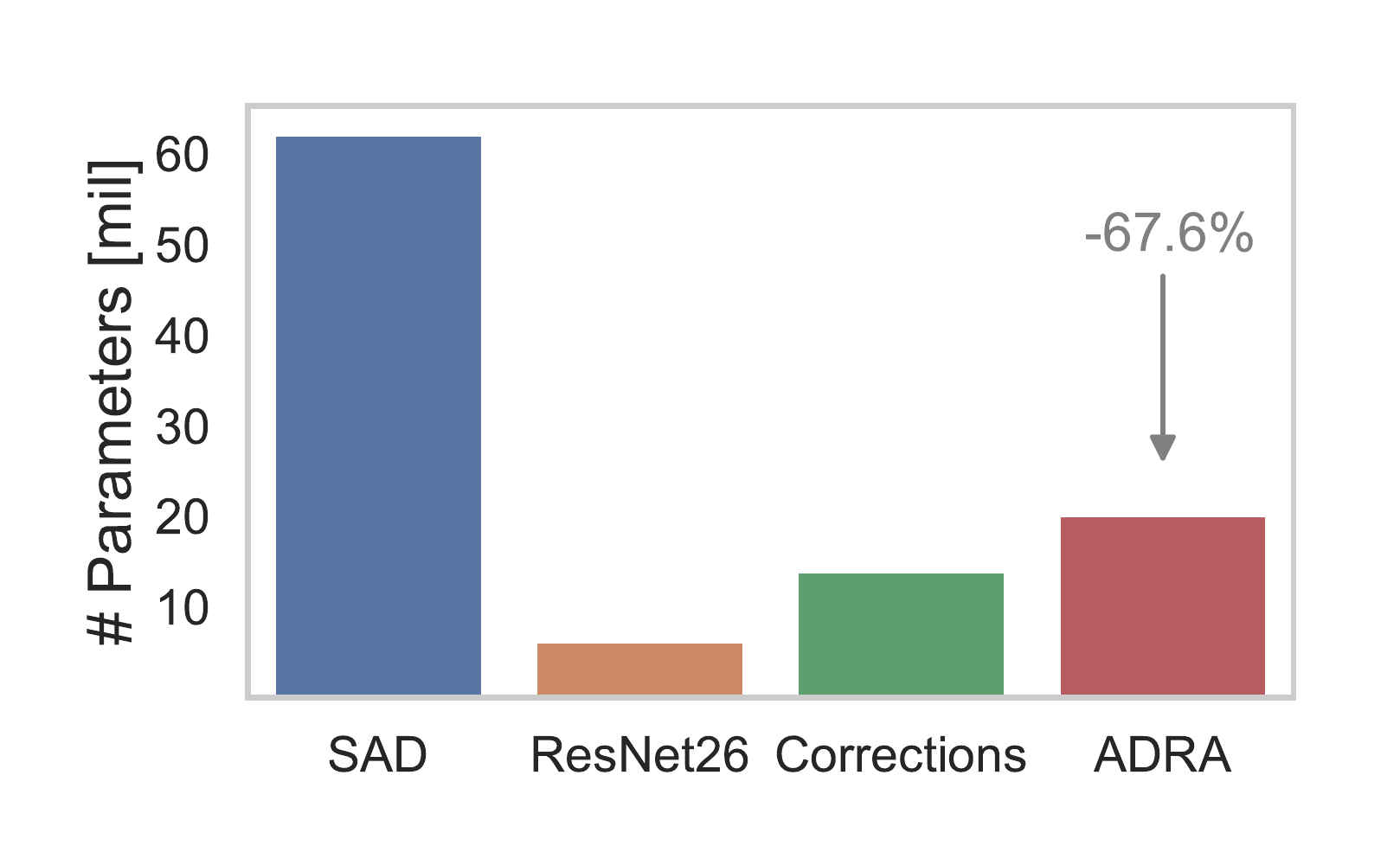}
\raisebox{0.15cm}{
\includegraphics[width=.47\textwidth]{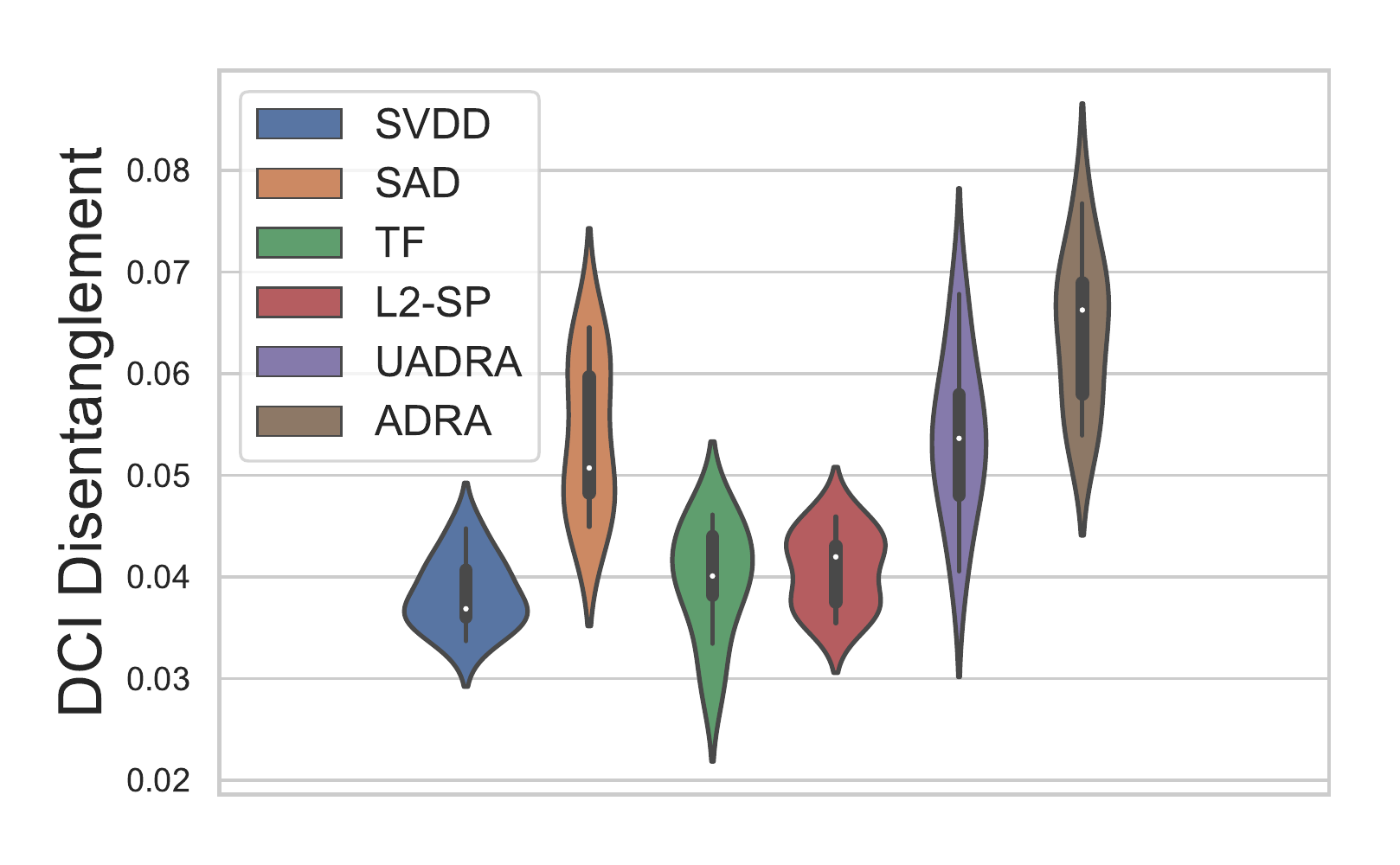}
}
}
\end{center}
\vspace*{-.5cm}
\caption{Left: memory requirements for incorporating ten tasks, just as in our benchmarks. \ours combines parameters of a single pretrained base model (\orangebullet) with task-specific corrections (\greenbullet), giving rise to a lean overall model (\redbullet) that incurs large savings compared to SAD (\bluebullet) and other previous methods. Calculation shown for $K=2$. Right: DCI disentanglement for different methods, both unsupervised (SVDD, UADRA) and using weak supervision through outlier exposure (TF, L\textsuperscript{2}-SP, SAD, \ours).} \label{fig:disentanglement}
\end{figure}

Next, we take a closer look at the representations learned by different anomaly detection methods. For this, we examine models on MPI3D \citep{gondal19}, a recently released dataset to facilitate research on disentangled representations. It contains joint pairs of ground-truth factors $z$ (color, shape, angle, \etc), and corresponding images $x$ of a robot arm mounted with an object. The original dataset comes in three different styles (photo-realistic, simple or detailed animation); we only make use of the more complex, photo-realistic images here.

We compare \ours to two fundamental approaches to deep anomaly detection: deep SVDD, a fully unsupervised one-class model proposed in \citep{ruff18}, and SAD \cite{ruff20a,ruff20b}, which can be viewed as a direct extension of SVDD that additionally incorporates outlier exposure into the learning criterion. Moreover, we include traditional finetuning (TF) and finetuning with regularization (L\textsuperscript{2}-SP \citep{li18}), both trained with outlier exposure. To ablate against weak supervision, we also train our method in an unsupervised fashion (UADRA), \ie with an empty corpus $Q_m=\emptyset$.

For our experiments on MPI3D, we arbitrarily fix a red cone as the normal object, and then train models on all available views. For methods that use outlier exposure, we include all such images that do not constitute the normal class. For example $z_\text{color}=\{ \text{white}, \text{green},  \text{blue}, \text{brown}, \text{olive} \}$ all appear in the corpus $Q_m$. For \ours, we can simply reuse the pretrained network also used in our CIFAR-10 benchmarks, and otherwise leave the optimization protocol unchanged from that in Section \ref{ssec:one-versus-rest}. To measure disentanglement, we follow \citet{locatello20} and evaluate models in terms of DCI disentanglement \citep{eastwood18}.

\paragraph{Results}

Disentanglement is often a desirable property \citep{bengio17,steenkiste19}. Representations learned via SVDD, \ie unsupervised anomaly detection, however exhibit relatively little of it (see Figure~\ref{fig:disentanglement}, right). Outlier exposure as in SAD raises disentanglement. This is in line with observations in \citet{locatello20}, which showed that some weak supervision is required for learning disentangled representations.

\ours exhibits the highest amount of disentanglement in its representations. To ensure that this actually stems from our model, we ablate this against UADRA. Removing access to the weak supervision of $Q_m$ causes another anticipated loss in disentanglement, but UADRA still disentangles better than SVDD. So while both our model and outlier exposure increase DCI disentanglement, their benefits should be viewed as distinct.

Interestingly TF and L\textsuperscript{2}-SP perform much worse than SAD, even though they only differ in their in parameter initialization. This is potentially because the pretrained network weights aren't \emph{particularly} well suited for the task, and the dataset size is too small in proportion to the number of free parameters to adapt them. ADRA on the other has a much smaller number of free parameters, thus allowing for sample-efficient utilization of those features which \emph{are} useful from the pretrained network.

\section{Conclusion}
\label{sec:conclusion}

Detecting anomalies is a difficult task, especially when carried out in high-dimensional spaces. In this paper, we introduced a powerful and simple method for deep anomaly detection. By incorporating dynamic residual adaptation to leverage pretrained models, \ours constitutes a parameter-efficient learning protocol.

Our method exhibits strong performance across common benchmarks for deep anomaly detection, and can robustly incorporate small modes of nominal data. Moreover, we established a positive link between the best performing methods for anomaly detection and their disentanglement, indicating that deep anomaly detection can directly benefit from the ongoing development of disentangled representations.

\bibliography{literature}

\begin{thebibliography}{79}
\providecommand{\natexlab}[1]{#1}
\providecommand{\url}[1]{\texttt{#1}}
\expandafter\ifx\csname urlstyle\endcsname\relax
  \providecommand{\doi}[1]{doi: #1}\else
  \providecommand{\doi}{doi: \begingroup \urlstyle{rm}\Url}\fi

\bibitem[Abati et~al.(2019)Abati, Porrello, Calderara, and Cucchiara]{abati19}
Davide Abati, Angelo Porrello, Simone Calderara, and Rita Cucchiara.
\newblock Latent space autoregression for novelty detection.
\newblock In \emph{IEEE Conference on Computer Vision and Pattern Recognition},
  pp.\  481--490, 2019.

\bibitem[Adhikari et~al.(2019)Adhikari, Ram, Tang, and Lin]{adhikari19}
Ashutosh Adhikari, Achyudh Ram, Raphael Tang, and Jimmy Lin.
\newblock Docbert: Bert for document classification.
\newblock \emph{arXiv preprint arXiv:1904.08398}, 2019.

\bibitem[Ahmed \& Courville(2019)Ahmed and Courville]{ahmed19}
Faruk Ahmed and Aaron Courville.
\newblock Detecting semantic anomalies.
\newblock \emph{arXiv preprint arXiv:1908.04388}, 2019.

\bibitem[Akcay et~al.(2018)Akcay, Atapour-Abarghouei, and Breckon]{akcay18}
Samet Akcay, Amir Atapour-Abarghouei, and Toby~P Breckon.
\newblock {GANomaly}: Semi-supervised anomaly detection via adversarial
  training.
\newblock In \emph{Asian Conference on Computer Vision}, pp.\  622--637.
  Springer, 2018.

\bibitem[Beltagy et~al.(2019)Beltagy, Lo, and Cohan]{beltagy19}
Iz~Beltagy, Kyle Lo, and Arman Cohan.
\newblock {SciBERT}: A pretrained language model for scientific text.
\newblock In \emph{Proceedings of the 2019 Conference on Empirical Methods in
  Natural Language Processing and the 9th International Joint Conference on
  Natural Language Processing}, pp.\  3606--3611, 2019.

\bibitem[Bengio(2017)]{bengio17}
Yoshua Bengio.
\newblock The consciousness prior.
\newblock \emph{arXiv preprint arXiv:1709.08568}, 2017.

\bibitem[Bergman \& Hoshen(2020)Bergman and Hoshen]{bergman20a}
Liron Bergman and Yedid Hoshen.
\newblock Classification-based anomaly detection for general data.
\newblock In \emph{International Conference on Learning Representations}, 2020.

\bibitem[Bergman et~al.(2020)Bergman, Cohen, and Hoshen]{bergman20b}
Liron Bergman, Niv Cohen, and Yedid Hoshen.
\newblock Deep nearest neighbor anomaly detection.
\newblock \emph{arXiv preprint arXiv:2002.10445}, 2020.

\bibitem[Bilen \& Vedaldi(2017)Bilen and Vedaldi]{bilen17}
Hakan Bilen and Andrea Vedaldi.
\newblock Universal representations: The missing link between faces, text,
  planktons, and cat breeds.
\newblock \emph{arXiv preprint arXiv:1701.07275}, 2017.

\bibitem[Bouchacourt et~al.(2018)Bouchacourt, Tomioka, and
  Nowozin]{bouchacourt18}
Diane Bouchacourt, Ryota Tomioka, and Sebastian Nowozin.
\newblock Multi-level variational autoencoder: Learning disentangled
  representations from grouped observations.
\newblock In \emph{Thirty-Second AAAI Conference on Artificial Intelligence},
  2018.

\bibitem[Burgess et~al.(2018)Burgess, Higgins, Pal, Matthey, Watters,
  Desjardins, and Lerchner]{burgess18}
Christopher~P Burgess, Irina Higgins, Arka Pal, Loic Matthey, Nick Watters,
  Guillaume Desjardins, and Alexander Lerchner.
\newblock Understanding disentangling in $\beta$-{VAE}.
\newblock \emph{arXiv preprint arXiv:1804.03599}, 2018.

\bibitem[Chandola et~al.(2009)Chandola, Banerjee, and Kumar]{chandola09}
Varun Chandola, Arindam Banerjee, and Vipin Kumar.
\newblock Anomaly detection: A survey.
\newblock \emph{ACM Computing Surveys (CSUR)}, 41\penalty0 (3):\penalty0 15,
  2009.

\bibitem[Chen et~al.(2018)Chen, Li, Grosse, and Duvenaud]{chen18}
Tian~Qi Chen, Xuechen Li, Roger~B Grosse, and David~K Duvenaud.
\newblock Isolating sources of disentanglement in variational autoencoders.
\newblock In \emph{Advances in Neural Information Processing Systems}, pp.\
  2610--2620, 2018.

\bibitem[Coates et~al.(2011)Coates, Ng, and Lee]{coates11}
Adam Coates, Andrew Ng, and Honglak Lee.
\newblock An analysis of single-layer networks in unsupervised feature
  learning.
\newblock In \emph{International Conference on Artificial Intelligence and
  Statistics}, pp.\  215--223, 2011.

\bibitem[Deecke et~al.(2018)Deecke, Vandermeulen, Ruff, Mandt, and
  Kloft]{deecke18}
Lucas Deecke, Robert Vandermeulen, Lukas Ruff, Stephan Mandt, and Marius Kloft.
\newblock Image anomaly detection with generative adversarial networks.
\newblock In \emph{Joint European Conference on Machine Learning and Knowledge
  Discovery in Databases}, pp.\  3--17. Springer, 2018.

\bibitem[Deecke et~al.(2020)Deecke, Timothy, and Bilen]{deecke20}
Lucas Deecke, Hospedales Timothy, and Hakan Bilen.
\newblock Latent domain learning with dynamic residual adapters.
\newblock \emph{arXiv preprint arXiv:2006.00996}, 2020.

\bibitem[Deng et~al.(2009)Deng, Dong, Socher, Li, Li, and Fei-Fei]{deng09}
Jia Deng, Wei Dong, Richard Socher, Li-Jia Li, Kai Li, and Li~Fei-Fei.
\newblock Imagenet: A large-scale hierarchical image database.
\newblock In \emph{IEEE Conference on Computer Vision and Pattern Recognition},
  pp.\  248--255, 2009.

\bibitem[Devlin et~al.(2018)Devlin, Chang, Lee, and Toutanova]{devlin18}
Jacob Devlin, Ming-Wei Chang, Kenton Lee, and Kristina Toutanova.
\newblock Bert: Pre-training of deep bidirectional transformers for language
  understanding.
\newblock \emph{arXiv preprint arXiv:1810.04805}, 2018.

\bibitem[Eastwood \& Williams(2018)Eastwood and Williams]{eastwood18}
Cian Eastwood and Christopher~KI Williams.
\newblock A framework for the quantitative evaluation of disentangled
  representations.
\newblock In \emph{International Conference on Learning Representations}, 2018.

\bibitem[Edgeworth(1887)]{edgeworth1887}
FY~Edgeworth.
\newblock {XLI}. on discordant observations.
\newblock \emph{The London, Edinburgh, and Dublin Philosophical Magazine and
  Journal of Science}, 23\penalty0 (143):\penalty0 364--375, 1887.

\bibitem[Emmott et~al.(2013)Emmott, Das, Dietterich, Fern, and Wong]{emmott13}
Andrew~F Emmott, Shubhomoy Das, Thomas Dietterich, Alan Fern, and Weng-Keen
  Wong.
\newblock Systematic construction of anomaly detection benchmarks from real
  data.
\newblock In \emph{ACM SIGKDD Workshop on Outlier Detection and Description},
  pp.\  16--21. ACM, 2013.

\bibitem[Erfani et~al.(2016)Erfani, Rajasegarar, Karunasekera, and
  Leckie]{erfani16}
Sarah~M Erfani, Sutharshan Rajasegarar, Shanika Karunasekera, and Christopher
  Leckie.
\newblock High-dimensional and large-scale anomaly detection using a linear
  one-class {SVM} with deep learning.
\newblock \emph{Pattern Recognition}, 58:\penalty0 121--134, 2016.

\bibitem[Gidaris et~al.(2018)Gidaris, Singh, and Komodakis]{gidaris18}
Spyros Gidaris, Praveer Singh, and Nikos Komodakis.
\newblock Unsupervised representation learning by predicting image rotations.
\newblock In \emph{International Conference on Learning Representations}, 2018.

\bibitem[Girshick(2015)]{girshick15}
Ross Girshick.
\newblock Fast {R-CNN}.
\newblock In \emph{IEEE Conference on Computer Vision and Pattern Recognition},
  pp.\  1440--1448, 2015.

\bibitem[Girshick et~al.(2014)Girshick, Donahue, Darrell, and
  Malik]{girshick14}
Ross Girshick, Jeff Donahue, Trevor Darrell, and Jitendra Malik.
\newblock Rich feature hierarchies for accurate object detection and semantic
  segmentation.
\newblock In \emph{IEEE Conference on Computer Vision and Pattern Recognition},
  pp.\  580--587, 2014.

\bibitem[Glorot \& Bengio(2010)Glorot and Bengio]{glorot10}
Xavier Glorot and Yoshua Bengio.
\newblock Understanding the difficulty of training deep feedforward neural
  networks.
\newblock In \emph{International Conference on Artificial Intelligence and
  Statistics}, pp.\  249--256, 2010.

\bibitem[Golan \& El-Yaniv(2018)Golan and El-Yaniv]{golan18}
Izhak Golan and Ran El-Yaniv.
\newblock Deep anomaly detection using geometric transformations.
\newblock In \emph{Advances in Neural Information Processing Systems}, pp.\
  9758--9769, 2018.

\bibitem[Gondal et~al.(2019)Gondal, Wuthrich, Miladinovic, Locatello, Breidt,
  Volchkov, Akpo, Bachem, Sch{\"o}lkopf, and Bauer]{gondal19}
Muhammad~Waleed Gondal, Manuel Wuthrich, Djordje Miladinovic, Francesco
  Locatello, Martin Breidt, Valentin Volchkov, Joel Akpo, Olivier Bachem,
  Bernhard Sch{\"o}lkopf, and Stefan Bauer.
\newblock On the transfer of inductive bias from simulation to the real world:
  a new disentanglement dataset.
\newblock In \emph{Advances in Neural Information Processing Systems}, pp.\
  15714--15725, 2019.

\bibitem[Grubbs(1969)]{grubbs1969}
Frank~E Grubbs.
\newblock Procedures for detecting outlying observations in samples.
\newblock \emph{Technometrics}, 11\penalty0 (1):\penalty0 1--21, 1969.

\bibitem[Guo et~al.(2019)Guo, Shi, Kumar, Grauman, Rosing, and Feris]{guo19}
Yunhui Guo, Honghui Shi, Abhishek Kumar, Kristen Grauman, Tajana Rosing, and
  Rogerio Feris.
\newblock Spottune: transfer learning through adaptive fine-tuning.
\newblock In \emph{IEEE Conference on Computer Vision and Pattern Recognition},
  pp.\  4805--4814, 2019.

\bibitem[He et~al.(2016)He, Zhang, Ren, and Sun]{he16}
Kaiming He, Xiangyu Zhang, Shaoqing Ren, and Jian Sun.
\newblock Deep residual learning for image recognition.
\newblock In \emph{IEEE Conference on Computer Vision and Pattern Recognition},
  pp.\  770--778, 2016.

\bibitem[He et~al.(2019)He, Girshick, and Doll{\'a}r]{he19}
Kaiming He, Ross Girshick, and Piotr Doll{\'a}r.
\newblock Rethinking imagenet pre-training.
\newblock In \emph{IEEE Conference on Computer Vision and Pattern Recognition},
  pp.\  4918--4927, 2019.

\bibitem[Hendrycks \& Gimpel(2017)Hendrycks and Gimpel]{hendrycks17}
Dan Hendrycks and Kevin Gimpel.
\newblock A baseline for detecting misclassified and out-of-distribution
  examples in neural networks.
\newblock In \emph{International Conference on Learning Representations}, 2017.

\bibitem[Hendrycks et~al.(2018)Hendrycks, Mazeika, and Dietterich]{hendrycks18}
Dan Hendrycks, Mantas Mazeika, and Thomas Dietterich.
\newblock Deep anomaly detection with outlier exposure.
\newblock \emph{arXiv preprint arXiv:1812.04606}, 2018.

\bibitem[Hendrycks et~al.(2019{\natexlab{a}})Hendrycks, Lee, and
  Mazeika]{hendrycks19a}
Dan Hendrycks, Kimin Lee, and Mantas Mazeika.
\newblock Using pre-training can improve model robustness and uncertainty.
\newblock \emph{arXiv preprint arXiv:1901.09960}, 2019{\natexlab{a}}.

\bibitem[Hendrycks et~al.(2019{\natexlab{b}})Hendrycks, Mazeika, Kadavath, and
  Song]{hendrycks19b}
Dan Hendrycks, Mantas Mazeika, Saurav Kadavath, and Dawn Song.
\newblock Using self-supervised learning can improve model robustness and
  uncertainty.
\newblock In \emph{Advances in Neural Information Processing Systems}, pp.\
  15637--15648, 2019{\natexlab{b}}.

\bibitem[Hendrycks et~al.(2020)Hendrycks, Liu, Wallace, Dziedzic, Krishnan, and
  Song]{hendrycks20}
Dan Hendrycks, Xiaoyuan Liu, Eric Wallace, Adam Dziedzic, Rishabh Krishnan, and
  Dawn Song.
\newblock Pretrained transformers improve out-of-distribution robustness.
\newblock \emph{arXiv preprint arXiv:2004.06100}, 2020.

\bibitem[Higgins et~al.(2017)Higgins, Matthey, Pal, Burgess, Glorot, Botvinick,
  Mohamed, and Lerchner]{higgins17}
Irina Higgins, Loic Matthey, Arka Pal, Christopher Burgess, Xavier Glorot,
  Matthew Botvinick, Shakir Mohamed, and Alexander Lerchner.
\newblock {beta-VAE}: Learning basic visual concepts with a constrained
  variational framework.
\newblock In \emph{International Conference on Learning Representations}, 2017.

\bibitem[Howard \& Ruder(2018)Howard and Ruder]{howard18}
Jeremy Howard and Sebastian Ruder.
\newblock Universal language model fine-tuning for text classification.
\newblock \emph{arXiv preprint arXiv:1801.06146}, 2018.

\bibitem[Kim \& Mnih(2018)Kim and Mnih]{kim18}
Hyunjik Kim and Andriy Mnih.
\newblock Disentangling by factorising.
\newblock In \emph{International Conference on Learning Representations}, 2018.

\bibitem[Kirkpatrick et~al.(2017)Kirkpatrick, Pascanu, Rabinowitz, Veness,
  Desjardins, Rusu, Milan, Quan, Ramalho, Grabska-Barwinska,
  et~al.]{kirkpatrick17}
James Kirkpatrick, Razvan Pascanu, Neil Rabinowitz, Joel Veness, Guillaume
  Desjardins, Andrei~A Rusu, Kieran Milan, John Quan, Tiago Ramalho, Agnieszka
  Grabska-Barwinska, et~al.
\newblock Overcoming catastrophic forgetting in neural networks.
\newblock \emph{Proceedings of the National Academy of Sciences}, 114\penalty0
  (13):\penalty0 3521--3526, 2017.

\bibitem[Krizhevsky \& Hinton(2009)Krizhevsky and Hinton]{krizhevsky09}
Alex Krizhevsky and Geoffrey Hinton.
\newblock Learning multiple layers of features from tiny images.
\newblock Technical report, University of Toronto, 2009.

\bibitem[Krizhevsky et~al.(2012)Krizhevsky, Sutskever, and
  Hinton]{krizhevsky12}
Alex Krizhevsky, Ilya Sutskever, and Geoffrey~E Hinton.
\newblock Imagenet classification with deep convolutional neural networks.
\newblock In \emph{Advances in Neural Information Processing Systems}, pp.\
  1097--1105, 2012.

\bibitem[Kulkarni et~al.(2015)Kulkarni, Whitney, Kohli, and
  Tenenbaum]{kulkarni15}
Tejas~D Kulkarni, William~F Whitney, Pushmeet Kohli, and Josh Tenenbaum.
\newblock Deep convolutional inverse graphics network.
\newblock In \emph{Advances in Neural Information Processing Systems}, pp.\
  2539--2547, 2015.

\bibitem[Kumar et~al.(2018)Kumar, Sattigeri, and Balakrishnan]{kumar18}
Abhishek Kumar, Prasanna Sattigeri, and Avinash Balakrishnan.
\newblock Variational inference of disentangled latent concepts from unlabeled
  observations.
\newblock In \emph{International Conference on Learning Representations}, 2018.

\bibitem[Li et~al.(2018)Li, Grandvalet, and Davoine]{li18}
Xuhong Li, Yves Grandvalet, and Franck Davoine.
\newblock Explicit inductive bias for transfer learning with convolutional
  networks.
\newblock In \emph{International Conference on Machine Learning}, pp.\
  2825--2834, 2018.

\bibitem[Liang et~al.(2018)Liang, Li, and Srikant]{liang18}
Shiyu Liang, Yixuan Li, and Rayadurgam Srikant.
\newblock Enhancing the reliability of out-of-distribution image detection in
  neural networks.
\newblock In \emph{International Conference on Learning Representations}, 2018.

\bibitem[Lin et~al.(2017)Lin, Feng, dos Santos, Yu, Xiang, Zhou, and
  Bengio]{lin17}
Zhouhan Lin, Minwei Feng, Cicero~Nogueira dos Santos, Mo~Yu, Bing Xiang, Bowen
  Zhou, and Yoshua Bengio.
\newblock A structured self-attentive sentence embedding.
\newblock In \emph{Proc. {ICLR}}, 2017.

\bibitem[Locatello et~al.(2019)Locatello, Bauer, Lucic, R{\"a}tsch, Gelly,
  Sch{\"o}lkopf, and Bachem]{locatello19}
Francesco Locatello, Stefan Bauer, Mario Lucic, Gunnar R{\"a}tsch, Sylvain
  Gelly, Bernhard Sch{\"o}lkopf, and Olivier Bachem.
\newblock Challenging common assumptions in the unsupervised learning of
  disentangled representations.
\newblock In \emph{International Conference on Machine Learning}, 2019.

\bibitem[Locatello et~al.(2020)Locatello, Poole, R{\"a}tsch, Sch{\"o}lkopf,
  Bachem, and Tschannen]{locatello20}
Francesco Locatello, Ben Poole, Gunnar R{\"a}tsch, Bernhard Sch{\"o}lkopf,
  Olivier Bachem, and Michael Tschannen.
\newblock Weakly-supervised disentanglement without compromises.
\newblock \emph{arXiv preprint arXiv:2002.02886}, 2020.

\bibitem[Lopez-Paz \& Ranzato(2017)Lopez-Paz and Ranzato]{lopezpaz17}
David Lopez-Paz and Marc'Aurelio Ranzato.
\newblock Gradient episodic memory for continual learning.
\newblock In \emph{Advances in Neural Information Processing Systems}, pp.\
  6467--6476, 2017.

\bibitem[Mahadevan et~al.(2010)Mahadevan, Li, Bhalodia, and
  Vasconcelos]{mahadevan10}
Vijay Mahadevan, Weixin Li, Viral Bhalodia, and Nuno Vasconcelos.
\newblock Anomaly detection in crowded scenes.
\newblock In \emph{IEEE Conference on Computer Vision and Pattern Recognition},
  pp.\  1975--1981, 2010.

\bibitem[Mikolov et~al.(2018)Mikolov, Grave, Bojanowski, Puhrsch, and
  Joulin]{mikolov18}
Tomas Mikolov, Edouard Grave, Piotr Bojanowski, Christian Puhrsch, and Armand
  Joulin.
\newblock Advances in pre-training distributed word representations.
\newblock In \emph{International Conference on Language Resources and
  Evaluation}, 2018.

\bibitem[Ourston et~al.(2003)Ourston, Matzner, Stump, and Hopkins]{ourston03}
Dirk Ourston, Sara Matzner, William Stump, and Bryan Hopkins.
\newblock Applications of hidden {Markov} models to detecting multi-stage
  network attacks.
\newblock In \emph{Proceedings of the 36th Annual Hawaii International
  Conference on System Sciences}. IEEE, 2003.

\bibitem[Paszke et~al.(2017)Paszke, Gross, Chintala, Chanan, Yang, DeVito, Lin,
  Desmaison, Antiga, and Lerer]{paszke17}
Adam Paszke, Sam Gross, Soumith Chintala, Gregory Chanan, Edward Yang, Zachary
  DeVito, Zeming Lin, Alban Desmaison, Luca Antiga, and Adam Lerer.
\newblock Automatic differentiation in {PyTorch}.
\newblock 2017.

\bibitem[Pelleg \& Moore(2005)Pelleg and Moore]{pelleg05}
Dan Pelleg and Andrew~W Moore.
\newblock Active learning for anomaly and rare-category detection.
\newblock In \emph{Advances in Neural Information Processing Systems}, pp.\
  1073--1080, 2005.

\bibitem[Perera et~al.(2019)Perera, Nallapati, and Xiang]{perera19}
Pramuditha Perera, Ramesh Nallapati, and Bing Xiang.
\newblock {OCGAN}: One-class novelty detection using gans with constrained
  latent representations.
\newblock In \emph{IEEE Conference on Computer Vision and Pattern Recognition},
  pp.\  2898--2906, 2019.

\bibitem[Pidhorskyi et~al.(2018)Pidhorskyi, Almohsen, and
  Doretto]{pidhorskyi2018}
Stanislav Pidhorskyi, Ranya Almohsen, and Gianfranco Doretto.
\newblock Generative probabilistic novelty detection with adversarial
  autoencoders.
\newblock In \emph{Advances in Neural Information Processing Systems}, pp.\
  6822--6833, 2018.

\bibitem[Rebuffi et~al.(2018)Rebuffi, Bilen, and Vedaldi]{rebuffi18}
S-A. Rebuffi, H.~Bilen, and A.~Vedaldi.
\newblock Efficient parametrization of multi-domain deep neural networks.
\newblock In \emph{IEEE Conference on Computer Vision and Pattern Recognition},
  2018.

\bibitem[Rebuffi et~al.(2017)Rebuffi, Bilen, and Vedaldi]{rebuffi17}
Sylvestre-Alvise Rebuffi, Hakan Bilen, and Andrea Vedaldi.
\newblock Learning multiple visual domains with residual adapters.
\newblock In \emph{Advances in Neural Information Processing Systems}, pp.\
  506--516, 2017.

\bibitem[Ruff et~al.(2018)Ruff, Vandermeulen, G{\"o}rnitz, Deecke, Siddiqui,
  Binder, M{\"u}ller, and Kloft]{ruff18}
Lukas Ruff, Robert Vandermeulen, Nico G{\"o}rnitz, Lucas Deecke, Shoaib~Ahmed
  Siddiqui, Alexander Binder, Emmanuel M{\"u}ller, and Marius Kloft.
\newblock Deep one-class classification.
\newblock In \emph{International Conference on Machine Learning}, pp.\
  4393--4402, 2018.

\bibitem[Ruff et~al.(2020{\natexlab{a}})Ruff, Vandermeulen, Franks, M{\"u}ller,
  and Kloft]{ruff20b}
Lukas Ruff, Robert~A Vandermeulen, Billy~Joe Franks, Klaus-Robert M{\"u}ller,
  and Marius Kloft.
\newblock Rethinking assumptions in deep anomaly detection.
\newblock \emph{arXiv preprint arXiv:2006.00339}, 2020{\natexlab{a}}.

\bibitem[Ruff et~al.(2020{\natexlab{b}})Ruff, Vandermeulen, G{\"o}rnitz,
  Binder, M{\"u}ller, M{\"u}ller, and Kloft]{ruff20a}
Lukas Ruff, Robert~A Vandermeulen, Nico G{\"o}rnitz, Alexander Binder, Emmanuel
  M{\"u}ller, Klaus-Robert M{\"u}ller, and Marius Kloft.
\newblock Deep semi-supervised anomaly detection.
\newblock In \emph{International Conference on Learning Representations},
  2020{\natexlab{b}}.

\bibitem[Sabokrou et~al.(2018)Sabokrou, Khalooei, Fathy, and
  Adeli]{sabokrou2018}
Mohammad Sabokrou, Mohammad Khalooei, Mahmood Fathy, and Ehsan Adeli.
\newblock Adversarially learned one-class classifier for novelty detection.
\newblock In \emph{IEEE Conference on Computer Vision and Pattern Recognition},
  pp.\  3379--3388, 2018.

\bibitem[Schlegl et~al.(2017)Schlegl, Seeb{\"o}ck, Waldstein, Schmidt-Erfurth,
  and Langs]{schlegl17}
Thomas Schlegl, Philipp Seeb{\"o}ck, Sebastian~M Waldstein, Ursula
  Schmidt-Erfurth, and Georg Langs.
\newblock Unsupervised anomaly detection with generative adversarial networks
  to guide marker discovery.
\newblock In \emph{International Conference on Information Processing in
  Medical Imaging}, pp.\  146--157. Springer, 2017.

\bibitem[Sch{\"o}lkopf \& Smola(2002)Sch{\"o}lkopf and Smola]{scholkopf02}
Bernhard Sch{\"o}lkopf and Alex~J Smola.
\newblock \emph{Learning with Kernels}.
\newblock MIT press, 2002.

\bibitem[Sch{\"o}lkopf et~al.(1999)Sch{\"o}lkopf, Platt, Shawe-Taylor, Smola,
  and Williamson]{schoelkopf99}
Bernhard Sch{\"o}lkopf, John~C Platt, John Shawe-Taylor, Alex~J Smola, and
  Robert~C Williamson.
\newblock Estimating the support of a high-dimensional distribution.
\newblock Technical Report MSR-TR-99-87, Microsoft Research, 1999.

\bibitem[Singliar \& Hauskrecht(2006)Singliar and Hauskrecht]{singliar06}
Tomas Singliar and Milos Hauskrecht.
\newblock Towards a learning traffic incident detection system.
\newblock In \emph{Workshop on Machine Learning Algorithms for Surveillance and
  Event Detection, International Conference on Machine Learning}, 2006.

\bibitem[Steinwart et~al.(2005)Steinwart, Hush, and Scovel]{steinwart05}
Ingo Steinwart, Don Hush, and Clint Scovel.
\newblock A classification framework for anomaly detection.
\newblock \emph{Journal of Machine Learning Research}, 6\penalty0
  (Feb):\penalty0 211--232, 2005.

\bibitem[Stickland \& Murray(2019)Stickland and Murray]{stickland19}
Asa~Cooper Stickland and Iain Murray.
\newblock {BERT} and {PALs}: Projected attention layers for efficient
  adaptation in multi-task learning.
\newblock In \emph{International Conference on Machine Learning}, 2019.

\bibitem[Sultani et~al.(2018)Sultani, Chen, and Shah]{sultani18}
Waqas Sultani, Chen Chen, and Mubarak Shah.
\newblock Real-world anomaly detection in surveillance videos.
\newblock In \emph{IEEE Conference on Computer Vision and Pattern Recognition},
  pp.\  6479--6488, 2018.

\bibitem[Torralba et~al.(2008)Torralba, Fergus, and Freeman]{torralba08}
Antonio Torralba, Rob Fergus, and William~T Freeman.
\newblock 80 million tiny images: A large data set for nonparametric object and
  scene recognition.
\newblock \emph{IEEE Transactions on Pattern Analysis and Machine
  Intelligence}, 30\penalty0 (11):\penalty0 1958--1970, 2008.

\bibitem[van Steenkiste et~al.(2019)van Steenkiste, Locatello, Schmidhuber, and
  Bachem]{steenkiste19}
Sjoerd van Steenkiste, Francesco Locatello, J{\"u}rgen Schmidhuber, and Olivier
  Bachem.
\newblock Are disentangled representations helpful for abstract visual
  reasoning?
\newblock In \emph{Advances in Neural Information Processing Systems}, pp.\
  14222--14235, 2019.

\bibitem[Yeung \& Chow(2002)Yeung and Chow]{yeung02}
Dit-Yan Yeung and Calvin Chow.
\newblock Parzen-window network intrusion detectors.
\newblock In \emph{International Conference on Pattern Recognition}, volume~4,
  pp.\  385--388. IEEE, 2002.

\bibitem[Zamir et~al.(2018)Zamir, Sax, Shen, Guibas, Malik, and
  Savarese]{zamir18}
Amir~R Zamir, Alexander Sax, William Shen, Leonidas~J Guibas, Jitendra Malik,
  and Silvio Savarese.
\newblock Taskonomy: Disentangling task transfer learning.
\newblock In \emph{IEEE Conference on Computer Vision and Pattern Recognition},
  pp.\  3712--3722, 2018.

\bibitem[Zhai et~al.(2016)Zhai, Cheng, Lu, and Zhang]{zhai2016}
Shuangfei Zhai, Yu~Cheng, Weining Lu, and Zhongfei Zhang.
\newblock Deep structured energy based models for anomaly detection.
\newblock In \emph{Proc. {ICML}}, volume~48, pp.\  1100--1109, 2016.

\bibitem[Zhou \& Paffenroth(2017)Zhou and Paffenroth]{zhou17}
Chong Zhou and Randy~C Paffenroth.
\newblock Anomaly detection with robust deep autoencoders.
\newblock In \emph{Proceedings of the 23rd ACM SIGKDD International Conference
  on Knowledge Discovery and Data Mining}, pp.\  665--674, 2017.

\bibitem[Zhou(2018)]{zhou18}
Zhi-Hua Zhou.
\newblock A brief introduction to weakly supervised learning.
\newblock \emph{National Science Review}, 5\penalty0 (1):\penalty0 44--53,
  2018.

\bibitem[Zong et~al.(2018)Zong, Song, Min, Cheng, Lumezanu, Cho, and
  Chen]{zong18}
Bo~Zong, Qi~Song, Martin~Renqiang Min, Wei Cheng, Cristian Lumezanu, Daeki Cho,
  and Haifeng Chen.
\newblock Deep autoencoding {Gaussian} mixture model for unsupervised anomaly
  detection.
\newblock In \emph{International Conference on Learning Representations}, 2018.

\end{thebibliography}
\bibliographystyle{iclr2019_conference}

\end{document}